%% file: imp.tex
\documentclass{article}

\usepackage{microtype}
\usepackage{graphicx}
\usepackage{booktabs}     
\usepackage{amsmath}      
\usepackage{amsfonts}     
\usepackage{bbm}          
\usepackage{float}        
\usepackage{subfig}       
\usepackage{wrapfig}      
\usepackage{nicefrac}     
\usepackage{gensymb}      
\usepackage{pbox}         
\usepackage{bm}           

\usepackage{hyperref}

\usepackage[accepted]{icml2019}

\icmltitlerunning{Infinite Mixture Prototypes}

\newcommand{\minisection}[1]{\textbf{#1}\hspace{0.3em}}

\DeclareMathOperator*{\argmax}{arg\,max}
\DeclareMathOperator*{\argmin}{arg\,min}

\renewcommand{\algorithmiccomment}[1]{\hfill$\triangleright$ #1}
\newcommand{\fullcomment}[1]{$\triangleright$ #1}

\begin{document}

\twocolumn[
\icmltitle{Infinite Mixture Prototypes for Few-Shot Learning}



\icmlsetsymbol{equal}{*}

\begin{icmlauthorlist}
\icmlauthor{Kelsey R. Allen}{mit}
\icmlauthor{Evan Shelhamer}{equal,berkeley}
\icmlauthor{Hanul Shin}{equal,mit}
\icmlauthor{Joshua B. Tenenbaum}{mit}
\end{icmlauthorlist}

\icmlaffiliation{mit}{Brain and Cognitive Sciences, MIT, Cambridge, MA}
\icmlaffiliation{berkeley}{Computer Science, UC Berkeley, Berkeley, CA}

\icmlcorrespondingauthor{Kelsey R. Allen}{krallen@mit.edu}

\icmlkeywords{meta-learning, metric learning, bayesian nonparametrics, few-shot learning}

\vskip 0.3in
]



\printAffiliationsAndNotice{\icmlEqualContribution} 

\begin{abstract}
We propose infinite mixture prototypes to adaptively represent both simple and complex data distributions for few-shot learning.
Our infinite mixture prototypes represent each class by a set of clusters, unlike existing prototypical methods that represent each class by a single cluster.
By inferring the number of clusters, infinite mixture prototypes interpolate between nearest neighbor and prototypical representations, which improves accuracy and robustness in the few-shot regime.
We show the importance of adaptive capacity for capturing complex data distributions such as alphabets, with 25$\%$ absolute accuracy improvements over prototypical networks, while still maintaining or improving accuracy on the standard Omniglot and mini-ImageNet benchmarks.
In clustering labeled and unlabeled data by the same clustering rule, infinite mixture prototypes achieves state-of-the-art semi-supervised accuracy.
As a further capability, we show that infinite mixture prototypes can perform purely unsupervised clustering, unlike existing prototypical methods.
\end{abstract}

\input{intro}
\input{method}
\input{results}
\input{rw}
\input{conclusion}

\section*{Acknowledgements}
We gratefully acknowledge support from DARPA grant 6938423 and KA is supported by NSERC. We thank Trevor Darrell and Ghassen Jerfel for advice and helpful discussions.

\bibliography{imp}
\bibliographystyle{icml2019}
\appendix
\input{appendix}

\end{document}

%% file: intro.tex
\section{Introduction}

Few-shot classification is the problem of learning to recognize new classes from only a few examples of each class \citep{lake2015human,fei2006one,miller2000learning}.
This requires careful attention to generalization, since overfitting or underfitting to the sparsely available data is more likely.
Nonparametric methods are well suited to this task, as they can model decision boundaries that more closely reflect the data distribution by using the data itself.

Two popular classes of nonparametric methods are nearest neighbor methods and prototypical methods. 
Nearest neighbor methods represent a class by storing all of its examples, and are high-capacity models that can capture complex distributions. 
Prototypical methods, such as Gaussian mixture models, represent a class by the mean of its examples, and are low-capacity models that can robustly fit simple distributions.
Neighbors and prototypes are thus two ends of a spectrum from complex to simple decision boundaries, and the choice of which to apply generally requires knowledge about the complexity of the distribution.

Adaptively modulating 
model capacity is thus an important problem, especially in few-shot learning where the complexity of individual tasks can differ.
Several approaches exist to tackle this, such as choosing $k$ for $k$-nearest neighbours, selecting the number of mixture components for Gaussian mixture models, or adjusting the bandwidth \citep{jones1996brief} for kernel density estimation \citep{parzen1962estimation}.

Infinite mixture modeling \citep{hjort2010bayesian} represents one way of unifying these approaches for adaptively setting capacity.
By inferring the number of mixture components for a given class from the data, it is possible to span the spectrum from nearest neighbors to prototypical representations.

This is particularly important in few-shot learning, where both underfitting and overfitting are common problems, because current models are fixed in their capacity.

\begin{figure}[t]
\vskip 0.2in
\centering
\includegraphics[width=\columnwidth]{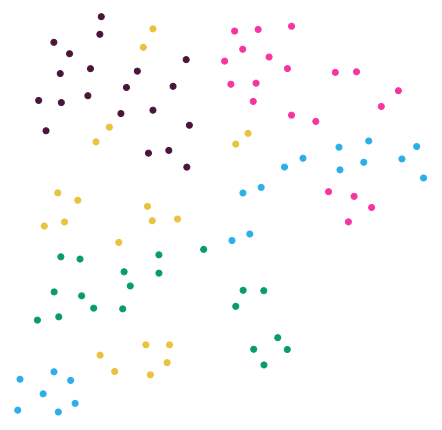}
\caption{
t-SNE visualization of the embedding from a prototypical network trained for alphabet recognition on Omniglot.
Each point is a character colored by its alphabet label.
The data distribution of each class is clearly not uni-modal, in violation of the modeling assumption for existing prototypical methods, causing errors.
Our infinite mixture prototypes represent each class by a \emph{set} of clusters, and infer their number, to better fit such distributions. 
}
\label{fig:embedding}
\vskip -0.2in
\end{figure}

To give an example, consider the problems of character and alphabet recognition.
Recognizing characters is fairly straightforward: each character looks alike, and can be
represented as a single prototype (a uni-modal Gaussian distribution).
Recognizing alphabets is more complex: the uni-modal distribution assumption could be violated, and a multi-modal approach could better capture the complexity of the distribution.
Figure \ref{fig:embedding} shows a prototypical network embedding for alphabets with this very issue.
Even though the embedding was optimized for uni-modality, the uni-modal assumption is not guaranteed on held-out data.

We therefore propose Infinite Mixture Prototypes (IMP) to represent a class as a set of clusters, with the number of clusters determined directly from the data.
IMP learns a deep embedding while also adapting the model capacity based on the complexity of the embedded data.
As a further benefit, the infinite mixture modeling approach can naturally incorporate unlabeled data.
We accordingly extend IMP to semi-supervised few-shot learning, and even to fully-unsupervised clustering inference.

An alternative approach to IMP would be to learn a parametric model.
The decision boundary would then be linear in the embedding, which is more complex than uni-modal prototypes, but less complex than nearest neighbors.
However, it may not be possible to find an embedding that yields a linear decision boundary.
In practice, either a parametric method or uni-modal mixture model is sensitive to the choice of model capacity, and may not successfully learn complex classes such as Omniglot \citep{lake2015human} alphabets.
Instead, a higher-capacity nonparametric method like nearest neighbors can work far better.
For simpler classes such as characters, a parametric model from a meta-learned initialization \citep{finn2017model} or a prototypical network that assumes uni-modal data \cite{snell2017prototypical} suffice.
Infinite mixture prototypes span these extremes, learning to adapt to both simple and complex classes.

In this paper, we extend prototypical networks from uni-modal to multi-modal clustering through infinite mixture modeling to give 25\% improvement in accuracy for alphabet recognition (complex classes) while preserving accuracy on character recognition (simple classes) on Omniglot.
In the semi-supervised setting infinite mixture prototypes are more accurate than semi-supervised prototypical networks. 
Infinite mixture modeling also allows for fully unsupervised clustering unlike existing prototypical methods.
We demonstrate that the DP-means algorithm is suitable for instantiating new clusters and that our novel extensions are necessary for best results in the few-shot regime.
By end-to-end learning with infinite mixture modeling, IMP adapts its model capacity to simple or complex data distributions, shown by equal or better accuracy compared to neighbors and uni-modal prototypes in all experiments.

\begin{figure*}[t]
\vskip 0.2in
\centering
\includegraphics[width=\textwidth]{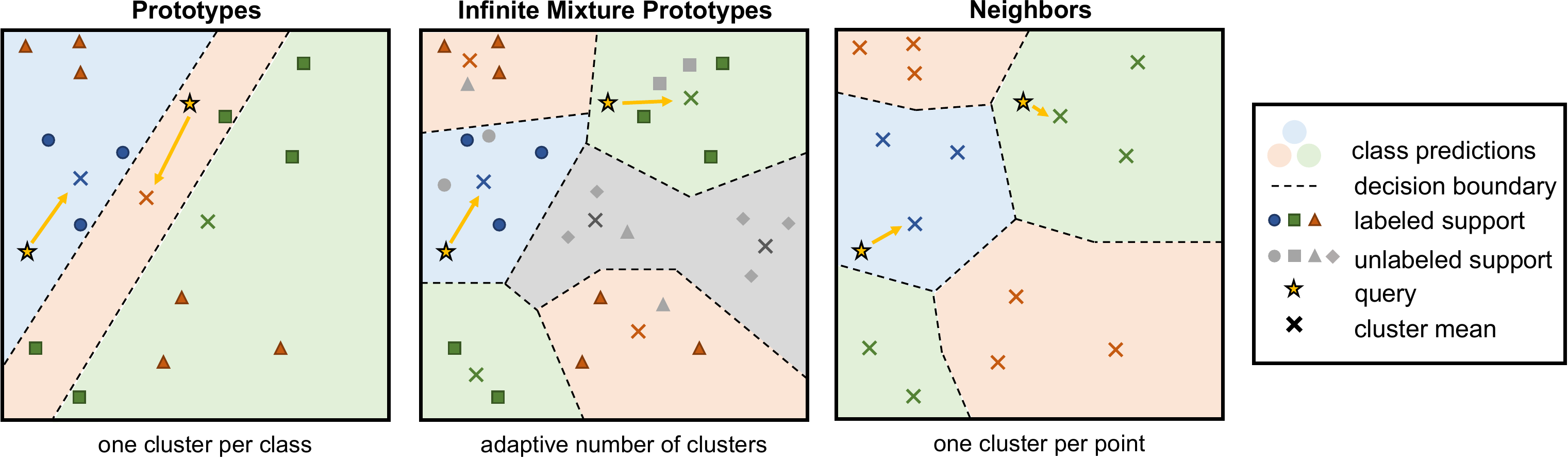}
\caption{
Our infinite mixture prototypes (IMP) method represents each class by a set of clusters, and infers the number of clusters from the data to adjust its modeling capacity.
IMP is optimized end-to-end to cluster labeled and unlabeled data into \emph{multi-modal} prototypes.
}
\label{fig:imp}
\vskip -0.2in
\end{figure*}

%% file: method.tex
\section{Background}
\label{sec:background}

For nonparametric representation learning methods, the model parameters are for the embedding function $h_{\phi} : \mathbb{R}^D \rightarrow \mathbb{R}^M$ that map an input point $x$ into a feature.
The embedding of point $x$ is the $M$-dimensional feature vector from the embedding function.
In deep models the parameters $\phi$ are the weights of a deep network, and the embedding is the output of the last layer of this network.
(Such methods are still nonparametric because they represent decisions by the embedding of the data, and not parameters alone.)

\subsection{Few-shot Classification}
\label{sec:few-shot}

In few-shot classification we are given a \emph{support} set $S = \{ (x_1, y_1), \ldots, (x_K, y_K) \}$ of $K$ labeled points and a \emph{query} set $Q = \{(x'_1,y'_1), \ldots, (x'_{K'},y'_{K'})\}$ of $K'$ labeled points where each $x_i, x'_i \in \mathbb{R}^D$ is a $D$-dimensional feature vector and $y_i, y'_i \in \{1, \ldots, N\}$ is the corresponding label.
In the semi-supervised setting, $y_i$ may not be provided for every point $x_i$.
The support set is for learning while the query set is for inference: the few-shot classification problem is to recognize the class of the queries given the labeled supports.

Few-shot classification is commonly learned by constructing few-shot tasks from a large dataset and optimizing the model parameters on these tasks.
Each task, comprised of the support and query sets, is called an \emph{episode}.
Episodes are drawn from a dataset by randomly sampling a subset of classes, sampling points from these classes, and then partitioning the points into supports and queries.
The number of classes in the support is referred to as the ``way'' of the episode, and the number of examples of each class is referred to as the ``shot'' of the episode.
Episodic optimization \citep{vinyals2016matching} iterates by making one episode and taking one update at a time.
The update to the model parameters is defined by the task loss, which for classification could be the softmax cross-entropy loss.

\subsection{Neighbors \& Prototypes}
\label{sec:neighbor-prototype}

\minisection{Neighbors}
Nearest neighbors classification \citep{cover1967nearest} assigns each query the label of the closest support.
Neighbor methods are extremely simple but remarkably effective, because the classification is local and so they can fit complex data distributions.
This generality comes at a computational cost, as the entire training set has to be stored and searched for inference.
More fundamentally, there is a modeling issue: how should the distance metric to determine the ``nearest'' neighbor be defined?

Neighborhood component analysis \citep{goldberger2004neighbourhood} learns the distance metric by defining stochastic neighbors to make the classification decision differentiable.
The metric is parameterized as a linear embedding $A$, and the probability of a point $x_i$ having neighbor $x_j$ is given by the softmax over Euclidean distances in the embedding:
\begin{equation}
  p_{ij} = \frac{\exp(\|Ax_i - Ax_j\|^2)}{\sum_{k \neq j} \exp(\|Ax_i - Ax_k\|^2)}.
\end{equation}
The probability that a point $x_i$ is in class $n$ is given by the sum of probabilities of neighbors in the class:
\begin{equation}
  p_A(y = n\,|\,x_i) = \sum_{j : y_j = n} p_{ij}.
\end{equation}

Stochastic neighbors naturally extend to a non-linear embedding trained by episodic optimization.
Deep nearest neighbors classification therefore serves as a high-capacity nonparametric method for few-shot learning.

\minisection{Prototypes}
Prototypical networks \citep{snell2017prototypical} form \emph{prototypes} as the mean of the embedded support points in each class:
\begin{equation}
  \mu_n = \frac{1}{|S_n|} \sum_{(x_i, y_i) \in S_n} h_{\phi}(x_i),
\end{equation}
with $S_n$ denoting the set of support points in class $n$.
Paired with a distance $d(x_i, x_j)$, the prototypes classify a query point $x'$ by the softmax over distances to the prototypes:
\begin{equation}
\label{eq:proto}
  p_{\phi}(y' = n\,|\,x') = \frac{\exp(-d(h_{\phi}(x'), \mu_n))}{\sum_{n'} \exp(-d(h_{\phi}(x'), \mu_{n'}))}.
\end{equation}
For the standard choice of the Euclidean distance function, the prototypes are equivalent to a Gaussian mixture model in the embedding with an identity covariance matrix.

$\phi$ is optimized by minimizing the negative log-probability of the true class of each query point by stochastic gradient descent over episodes.
Prototypical networks therefore learn to create \emph{uni-modal} class distributions for \emph{fully-labeled} supports by representing each class by one cluster.

\subsection{Infinite Mixture Modeling}
\label{sec:infinite-mixture}

Infinite mixture models \citep{hjort2010bayesian} do not require the number of mixture components to be known and finite.
Instead, the number of components is inferred from data through Bayesian nonparametric methods \citep{west1994hierarchical,rasmussen2000infinite}.
In this way infinite mixture models adapt their capacity to steer between overfitting with high capacity and underfitting with low capacity.

The advantage of adaptivity is countered by the implementation and computational difficulties of Gibbs sampling and variational inference for infinite mixtures.
To counter these issues, DP-means \citep{kulis2011revisiting} is a deterministic, hard clustering algorithm derived via Bayesian nonparametrics for the Dirichlet process.
DP-means iterates over the data points, computing each point's minimum distance to all existing cluster means.
If this distance is greater than a threshold $\lambda$, a new cluster is created with mean equal to the point.
It optimizes a $k$-means-like objective for reconstruction error plus a penalty for making clusters.

$\lambda$, the distance threshold for creating a new cluster, is the sole hyperparameter for the algorithm.
In deriving DP-means, \citet{kulis2011revisiting} relate $\alpha$, the concentration parameter for the Chinese restaurant process (CRP) \citep{aldous1985exchangeability}, to $\lambda$: 
\begin{equation}
\label{eq:lambda}
\lambda = 2\sigma\log(\frac{\alpha}{(1+\frac{\rho}{\sigma})^{d/2}})
\end{equation}
where $\rho$ is a measure of the standard deviation for the base distribution from which clusters are assumed to be drawn in the CRP. 
They then derive DP-means by connection to a Gibbs sampling procedure in the limit as $\sigma$ approaches 0.

\section{Infinite Mixture Prototypes (IMP)}
\label{sec:imp}
Our infinite mixture prototypes (IMP) method pursues two approaches for adapting capacity: learning cluster variance to scale assignments, and multi-modal clustering to interpolate between neighbor and prototypical representations. 
This capability lets our model adapt its capacity to avoid underfitting, unlike existing prototypical models with fixed capacity.
Figure \ref{fig:imp} gives a schematic view of our multi-modal representation and how it differs from existing prototype and neighbor representations.
Algorithm \ref{alg:imp} expresses infinite mixture prototypes inference in pseudocode.

Within an episode, we initially cluster the support into class-wise means.
Inference proceeds by iterating through all support points and computing their minimum distance to all existing clusters.
If this distance exceeds a threshold $\lambda$, a new cluster is made with mean equal to that point.
IMP then updates soft cluster assignments $z_{i,c}$ as the normalized Gaussian density for cluster membership.
Finally, cluster means $\mu_c$ are computed by the weighted mean of their members.
Since each class can have multiple clusters, we classify a query point $x'$ by the softmax over distances to the closest cluster in each class $n$:
\begin{equation}
\label{eq:imp-query}
p_{\phi}(y' = n\,|\,x') = \frac{\exp(-d(h_{\phi}(x'),\: \mu_{c^*_n}))}{\sum_{n'} \exp(-d(h_{\phi}(x'), \: \mu_{c^*_{n'}}))}
\end{equation}
with $c^*_n = \argmin_{c: l_c = n}d(h_\phi(x'), \mu_c)$ indexing the clusters, where each cluster $c$ has label $l_c$.

IMP optimizes the embedding parameters $\phi$ and cluster variances $\sigma$ by stochastic gradient descent across episodes. 

\subsection{Adapting capacity by learning cluster variance $\sigma$}
\label{sec:learning-variance}

We learn the cluster variance $\sigma$ to scale the assignment of support points to clusters.
When $\sigma$ is small, the effective distance is large and the closest points dominate, and when $\sigma$ is large, the effective distance is small so farther points are more included.
$\sigma$ is differentiable, and therefore learned jointly with the embedding parameters $\phi$.
In practice, learning $\sigma$ can improve the accuracy of prototypical networks, which we demonstrate by ablation in Table \ref{tab:ablate-sigma-modes}.
For IMP, $\sigma$ has a further role in creating new clusters. 

\subsection{Adapting capacity by multi-modal clustering}
\label{sec:e2e-modes}
\begin{algorithm}[t]
\small
\caption{IMP: support prototypes and query inference}
\label{alg:imp}
\textbf{Require:} supports $(x_1,y_1) ..., (x_K,y_K)$ and queries $x'_1, ..., x'_{K'}$
\textbf{Return:} clusters $(\mu_c, l_c, \sigma_c)$ and query classifications $p(y' | x')$

\centering
\begin{algorithmic}
\STATE 1. Init. each cluster $\mu_c$ with label $l_c$  and $\sigma_c = \sigma_l$ as class-\\
\quad wise means of the supports, and $C$ as the number of classes
\STATE 2. Estimate $\lambda$ as in Equation \ref{eq:lambda}
\STATE 3. Infer the number of clusters
\FOR{each point ${x}_i$}
\FOR{$c$ in $\{1,...,C\}$}
\STATE $d_{i,c} = \begin{cases}
    \|h_{\phi}({x}_i) - {\mu}_c\|^2 & \text{if (${x}_i$ is labeled and $l_c = y_i)$}\\& \text{or ${x}_i$ is unlabeled} \\
    +\infty & \text{otherwise } \\
    \end{cases}$
\ENDFOR
\STATE If $\min_c d_{ic} > \lambda$: set $C = C+1,$ ${\mu}_C = h_{\phi}({x}_i)$, $l_C = y_i$, $\sigma_C = \{ \sigma_l$ if $x_i$ labeled, $\sigma_u$ otherwise$\}$.
\ENDFOR
\STATE 4. Assign supports to clusters by $z_{i,c} = \frac{\mathcal{N}(h_{\phi}(x_i);\mu_c, \sigma_c)}{\sum_c \mathcal{N}(h_{\phi}(x_i);\mu_c,\sigma_c)}$ 
\STATE 5. For each cluster $c$, compute mean $\mu_c = \frac{\sum_{i}{z_{i,c}h_{\phi}(x_i)}}{\sum_i{z_{i,c}}}$ \\
\STATE 6. Classify queries by Equation \ref{eq:imp-query}
\end{algorithmic}
\end{algorithm}

To create multi-modal prototypes, we extend the clustering algorithm DP-means \citep{kulis2011revisiting} for compatibility with classification and end-to-end optimization.
For classification, we distinguish labeled and unlabeled clusters, and incorporate labels into the point-cluster distance calculation.
For end-to-end optimization, we soften cluster assignment, propose a scheme to select $\lambda$, and mask the classification loss to encourage multi-modality. 

\minisection{Indirect optimization of $\lambda$}
While $\lambda$ is non-differentiable, we propose an indirect optimization of the effective threshold for creating a new cluster.
Based on Equation \ref{eq:lambda}, $\lambda$ depends on the concentration hyperparameter $\alpha$, a measure of standard deviation in the prior $\rho$, and the cluster variance $\sigma$.
$\alpha$ remains a hyperparameter, but with lessened effect.
We estimate $\rho$ as the variance between prototypes in each episode.
As noted, $\sigma$ is differentiable, so we learn it.

We model separate variances for labeled and unlabeled clusters, $\sigma_l$ and $\sigma_u$ respectively, in order to capture differences in uncertainty between labeled and unlabeled data.
In the fully-supervised setting, $\lambda$ is estimated from $\sigma_l$, and in the semi-supervised setting $\lambda$ is estimated from the mean of $\sigma_l$ and $\sigma_u$.
In summary, learning the cluster variances $\sigma$ affects IMP by scaling the distances between points and clusters, and through its role in our episodic estimation of $\lambda$.

\minisection{Multi-modal loss} 
We optimize all models with the cross-entropy loss.
For the multi-modal methods (nearest neighbors and IMP), we mask the loss to only include the closest neighbor/cluster for each class, in the same manner as inference.
That is, for a class $n$, we find the most likely cluster $c^*_n \gets \argmax_{c: l_c = n} \log p(h_\phi(x)|\mu_c, \sigma_c)$ and then take the loss over the queries in the class ($Q_n$):
\begin{equation*}
\begin{aligned}
J &= \frac{1}{|Q_n|}\sum_{x \in Q_n}\bigg[-\log p(h_\phi(x)\:|\:\mu_{c^*_n}, \sigma_{c^*_n}) \: + \\
  &\log\sum_{n'\neq n} p(h_\phi(x)\:|\:\mu_{c^*_{n'}}, \sigma_{c^*_{n'}})\bigg]. \\
\end{aligned}
\end{equation*}
Taking the loss for the closest clusters avoids over-penalizing multi-modality in the embedding, as taking the loss over all the clusters would.
We found that masking improves the few-shot accuracy of these methods over other losses that incorporate all clusters.

\begin{table}[t]
\caption{
Multi-modal clustering and learning cluster variances on fully-supervised 10-way, 10-shot Omniglot alphabet recognition and 5-way, 5-shot mini-ImageNet.
Scaling distances with the learned variance gives a small improvement and multi-modal clustering gives a further improvement.
}
\label{tab:ablate-sigma-modes}
\vskip 0.15in
\begin{center}
\begin{small}
\begin{sc}
\begin{tabular}{lcccc}
  \toprule
  method & $\sigma$ & \pbox{5em}{multi-\\modal} & alph. acc. & mini. acc. \\
  \midrule
  prototypes & - & - & 65.2 $\pm$ 0.6 & 66.1 $\pm$ 0.6 \\
  prototypes & \checkmark & - & 65.2 $\pm$ 0.6 & 67.2 $\pm$ 0.5 \\
  IMP (ours) & \checkmark & \checkmark & \textbf{92.0} $\pm$ 0.1 & \textbf{68.1} $\pm$ 0.8 \\
  \bottomrule
\end{tabular}
\end{sc}
\end{small}
\end{center}
\vskip -0.1in
\end{table}

\begin{table}[t]
\caption{
Learning labeled cluster variance $\sigma_l$ and unlabeled cluster variance $\sigma_u$ on semi-supervised 5-way, 1-shot Omniglot and mini-ImageNet with 5 unlabeled points per class and 5 distractors (see Section \ref{sec:results}).
Learning $\sigma_l, \sigma_u$ is better than learning a tied $\sigma$ for labeled and unlabeled clusters.
}
\label{tab:ablate-tied-sigmas}
\vskip 0.15in
\begin{center}
\begin{small}
\begin{sc}
\begin{tabular}{lccc}
\toprule
method & $\sigma$ & omni. acc. & mini. acc. \\
\midrule
tied & $\sigma$ & 93.5$\pm$0.3 & 48.6$\pm$0.4 \\
imp (ours) & $\sigma_l, \sigma_u$ & \textbf{98.9}$\pm$0.1 & \textbf{49.6}$\pm$0.8 \\
\bottomrule
\end{tabular}
\end{sc}
\end{small}
\end{center}
\vskip -0.1in
\end{table}

\begin{figure}[t]
\vskip 0.2in
\centering
\includegraphics[width=\columnwidth]{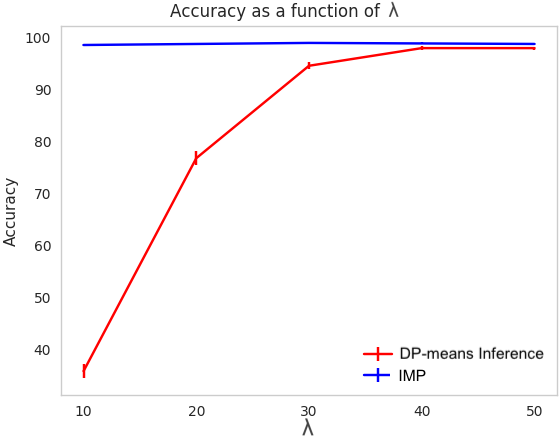}
\caption{
Learning and inference with IMP is more accurate and robust than DP-means inference on a prototypical network embedding alone.
This plot shows the accuracy for the standard benchmark of semi-supervised 5-way, 1-shot Omniglot for different choices of the distance threshold $\lambda$ for creating a new cluster.
}
\label{fig:ablate-lambda-sigma}
\vskip -0.2in
\end{figure}

\subsection{Ablations and Alternatives}
We ablate our episodic and end-to-end extensions of DP-means to validate their importance for few-shot learning.
Learning and performing inference with IMP is more robust to different choices of $\lambda$ than simply using DP-means during inference (Figure \ref{fig:ablate-lambda-sigma}). 
Multi-modality and learned variance make their own contributions to accuracy (Table \ref{tab:ablate-sigma-modes}).
Learning separate $\sigma_l, \sigma_u$, for labeled and unlabeled clusters respectively, is more accurate than learning a shared $\sigma$ for all clusters (Table \ref{tab:ablate-tied-sigmas}).
For full details of the datasets and settings in these ablations, refer to Section \ref{sec:results}.

In principle, IMP's clustering can be iterated multiple times during training and inference. 
However, we found that one clustering iteration suffices.
Two iterations during training had no effect on accuracy, and even 100 iterations during inference still had no effect on accuracy, showing that the clustering is stable.

\minisection{Alternative Algorithms} 
DP-means was derived through the limit of a Gibbs sampler as the variance approaches 0, and so it does hard assignment of points to clusters. 
With hard assignment, it is still possible to learn the embedding parameters $\phi$ end-to-end by differentiating through the softmax over distances between query points and support clusters as in Equation \ref{eq:proto}.
However, hard assignment of labeled and unlabeled data is harmful in our experiments, especially early on in training (see supplement).

When reintroducing variance into multi-modal clustering as we do, a natural approach would be to reconsider the Gibbs sampler for the CRP \citep{west1994hierarchical,neal2000markov} from which DP-means was derived, or other Dirichlet process inference methods such as expectation maximization \citep{kimura2013expectation}.
These alternatives are less accurate in our experiments, mainly as a result of the CRP prior's ``rich get richer'' dynamics, which prefers clusters with more assignments (leading to accuracy drops of 5--10$\%$).
This is especially problematic early in training, when unlabeled points are often incorrectly assigned.
The supplement includes derivations and experiments regarding these multi-modal clustering alternatives.

%% file: results.tex
\section{Experiments}
\label{sec:results}

\begin{table*}[t]
\caption{
Alphabet and character recognition accuracy on Omniglot.
Alphabets have more complex, multi-modal data distributions while characters have simpler, uni-modal data distributions.
IMP improves accuracy for multi-modal alphabet classes, preserves accuracy for uni-modal character classes (Chars), and generalizes better from super-classes to sub-classes.
}
\label{tab:alpha-char}
\vskip 0.15in
\begin{center}
\begin{small}
\begin{sc}
\begin{tabular}{llllll}
  \toprule
  Training    &  Testing & Prototypes & IMP & Neighbors \\  
  \midrule
  Alphabet & Alphabet (10-way 10-shot) & 65.6$\pm$0.4 & 92.0$\pm$0.1 & \textbf{92.4}$\pm$0.2 \\ 
  Alphabet & Chars (20-way 1-shot) & 82.1$\pm$0.4 & \textbf{95.4}$\pm$0.2 & \textbf{95.4}$\pm$0.2 \\ 
  Chars & Chars (20-way 1-shot) & \textbf{94.9}$\pm$0.2 & \textbf{95.1}$\pm$0.1& \textbf{95.1}$\pm$0.1 \\ 
  \bottomrule
\end{tabular}
\end{sc}
\end{small}
\end{center}
\vskip -0.1in
\end{table*}
We experimentally show that infinite mixture prototypes are more accurate and more general than uni-modal prototypes.

We control for architecture and optimization by comparing methods with the same base architecture of \citet{vinyals2016matching} and same episodic optimization settings of \citet{snell2017prototypical}.
For further implementation details see Appendix A.1 of the supplement.
All code for our method and baselines will be released.

We consider two widely-used datasets for few-shot learning:

\minisection{Omniglot} \citep{lake2015human} is a dataset of 1,623 handwritten characters from 50 alphabets.
There are 20 examples of each character, where the images are resized to 28x28 pixels and each image is rotated by multiples of 90$\degree$.
This gives 6,492 classes in total, which are then split into 4,112 training classes, 688 validation classes, and 1,692 test classes.

\minisection{mini-ImageNet} \citep{vinyals2016matching} is a reduced version of the ILSVRC'12 dataset \citep{russakovsky2015imagenet}, which contains 600 84x84 images for 100 classes randomly selected from the full dataset.
We use the split from \citet{ravi2016optimization} with 64/16/20 classes for train/val/test.

\subsection{Accuracy and Generality of Multi-modal Clustering by Infinite Mixture Prototypes}

Our experiments on Omniglot alphabets and characters show that multi-modal prototypes are significantly more accurate than uni-modal prototypes for recognizing complex classes (alphabets) and recover uni-modal prototypes as a special case for recognizing simple classes (characters).
Multi-modal prototypes generalize better for super-class to sub-class transfer learning, improving accuracy when training on alphabets but testing on characters.
By unifying the clustering of labeled and unlabeled data alike, our multi-modal prototypes also address fully unsupervised clustering, unlike prior prototypical network models that are undefined without labels.

We first show the importance of multi-modality for learning representations of multi-modal classes: Omniglot alphabets.
For these experiments, we train for alphabet classification, using only the super-class labels.
Episodes are constructed by sampling $n$ alphabets, and $n_c$ characters within each alphabet. 1 image of each character is randomly sampled for the support, with 5 examples of each character for the query. 
We refer to these episodes as $n$-way $n_c$-shot episodes.
For training, we sample 10-way, 10-shot episodes.

For character testing, we provide 1 labeled image of 20 different characters in the support, and score the correct character classification of the queries.
Note that both alphabet and character testing are on held-out alphabets and characters respectively.

As seen in Table \ref{tab:alpha-char}, IMP substantially outperforms prototypical networks for both alphabet and character recognition from alphabet training.
On 20-way 1-shot character recognition, IMP achieves $95.4\%$ from alphabet supervision alone, slightly out-performing prototypical networks trained directly on character recognition ($94.9\%$).
By clustering each super-class into multiple modes, IMP is better able to generalize to sub-classes.

For a parametric alternative, we trained MAML \citep{finn2017model} on alphabet recognition, with the same episode composition as IMP. 
MAML achieves only 61.9$\%$ accuracy on 10-way 10-shot alphabet recognition. 
This demonstrates that a parametric classifier of this capacity, with decisions that are linear in the embedding, is not enough to solve alphabet recognition---instead, multi-modality is necessary.

\begin{table}[h]
\caption{
Generalization to held-out characters on 10-way, 5-shot Omniglot alphabet recognition.
$40\%$ of the characters are kept for training and $60\%$ held out for testing.
IMP maintains accuracy on held-out characters, suggesting that multi-modal clustering is more robust to new and different sub-classes from the same super-class.}
\label{tab:alpha-modes}
\vskip 0.15in
\begin{center}
\begin{small}
\begin{sc}
\begin{tabular}{llll}
  \toprule
  Method    &  \pbox{5em}{Training\\Modes} & \pbox{5em}{Testing\\Modes} & \pbox{5em}{Both\\Modes} \\
  \midrule
   IMP (ours) & \textbf{99.0}$\pm$0.1 & \textbf{94.9}$\pm$0.2 & \textbf{96.6}$\pm$ 0.2 \\
   Prototypes & 92.4$\pm$0.3 & 77.7$\pm$0.4 & 82.9$\pm$0.4\\
  \bottomrule
\end{tabular}
\end{sc}
\end{small}
\end{center}
\vskip -0.1in
\end{table}

To further examine generalization, we consider holding out character sub-classes during alphabet super-class training.
In this experiment the training and testing alphabets are the same, but the characters within each alphabet are divided into training (40\%) and testing (60\%) splits.
We compare alphabet recognition accuracy using training characters, testing characters, and all characters to measure generalization to held-out modes in Table \ref{tab:alpha-modes}.
While prototypical networks achieve good accuracy on training modes, their accuracy drops 16\% relative on testing modes, and still drops 10\% relative on the combination of both modes.
The reduced accuracy of prototypical networks on held-out modes indicates that uni-modality is not maintained on unseen characters even when they are from the same alphabets.
IMP accuracy drops less than 5\% relative from training to testing modes and both modes, showing that multi-modal clustering generalizes better to unseen data.

\minisection{Fully Unsupervised Clustering}
IMP is able to do fully unsupervised clustering via multi-modality.
Prototypical networks \citep{snell2017prototypical} and semi-supervised prototypical networks \citep{ren18fewshotssl} are undefined without labeled data during testing because the number of clusters is defined by the number of classes.

\begin{table}[t]
\caption{
Unsupervised clustering of unseen Omniglot characters by IMP.
Learning with IMP makes substantially purer clusters than DP-means inference on a prototypical network embedding,
showing that the full method is necessary for best results.
}
\label{tab:unsup-scores}
\vskip 0.05in
\begin{center}
\begin{small}
\begin{sc}
\begin{tabular}{llllll}
\toprule
Method & Metric & 10-way & 100-way & 200-way \\
\midrule
IMP & Purity & \textbf{0.97} & \textbf{0.90} & \textbf{0.91} \\
DP-means &  & 0.91 & 0.73 & 0.71\\
IMP & NMI & \textbf{0.97} & \textbf{0.95} & \textbf{0.94} \\
DP-means &  & 0.89 & 0.88 & 0.87\\
IMP & AMI & \textbf{0.92} & \textbf{0.81} & \textbf{0.70} \\
DP-means & & 0.76 & 0.58 & 0.51 \\
\bottomrule
\end{tabular}
\end{sc}
\end{small}
\end{center}
\vskip -0.3in
\end{table}

For this unsupervised clustering setting, we use the models that were optimized for alphabet recognition.
For testing, we randomly sample 5 examples of $n$ character classes from the test set without labels.

IMP handles labeled and unlabeled data by the same clustering rule, infers the number of clusters as needed, and achieves good results under the standard clustering metrics of purity, and normalized/adjusted mutual information (NMI/AMI).
We examine IMP's clustering quality on purely unlabeled data in Table \ref{tab:unsup-scores}. 
IMP maintains strong performance across a large number of unlabeled clusters, without knowing the number of classes in advance, and without having seen any examples from the classes during training.

As a baseline, we evaluate multi-modal inference by DP-means \citep{kulis2011revisiting} on the embedding from a prototypical network with the same architecture and training data as IMP.
We cross-validate the cluster threshold $\lambda$ on validation episodes for each setting, choosing by AMI.

\subsection{Few-Shot Classification Benchmarks}
\label{sec:few-shot-results}

We evaluate IMP on the standard few-shot classification benchmarks of Omniglot and mini-ImageNet in the fully-supervised and semi-supervised regimes.

We consider five strong fully-supervised baselines trained on 100$\%$ of the data.
We compare to three parametric methods, MAML \citep{finn2017model}, Reptile \citep{nichol2018reptile}, and few-shot graph networks \citep{garcia2017few}, as well as three nonparametric methods, nearest neighbors, prototypical networks \cite{snell2017prototypical}, and the memory-based model of \citet{KaiserNRB17}.

Fully-supervised results are reported in Table \ref{tab:full}.
In this setting, we evaluate IMP in the standard episodic protocol of few-shot learning: shot and way are fixed and classes are balanced within an episode.
IMP learns to recover uni-modal clustering as a special case, matching or out-performing prototypical networks when the classes are uni-modal.

In the semi-supervised setting of labeled and unlabeled examples we follow \citet{ren18fewshotssl}.
We take only 40$\%$ of the data as labeled \emph{for both supports and queries} while the rest of the data is included as unlabeled examples.
The unlabeled data is then incorporated into episodes as (1) within-support examples that allow for semi-supervised refinement of the support classes or (2) \emph{distractors} which lie in the complement of the support classes.
Semi-supervised episodes augment the fully supervised $n$-way, $k$-shot support with $5$ unlabeled examples for each of the $n$ classes and include $5$ more distractor classes with $5$ unlabeled instances each.
The query set still contains only support classes.

Semi-supervised results are reported in Table \ref{tab:semi}.
We train and test IMP, existing prototypical methods, and nearest neighbors in this setting.
Semi-supervised prototypical networks \citep{ren18fewshotssl} incorporate unlabeled data by soft $k$-means clustering (of their three comparable variants, we report ``Soft $k$-Means+Cluster'' results).  
Prototypical networks \citep{snell2017prototypical} and neighbors are simply trained on the $40\%$ of the data with labels.

Through multi-modality, IMP clusters labeled and unlabeled data by a single rule.
In particular this helps with the distractor distribution, which is in fact more diffuse and multi-modal by comprising several different classes.

The results reported on these benchmarks are for models trained and tested with $n$-way episodes.
This is to equalize comparison across methods\footnote{
\citet{snell2017prototypical} train at higher way than testing and report a boost in accuracy.
We find that this boost is somewhat illusory, and at least partially explained away by controlling for the number of gradients per update.
We show this by experiment through the use of gradient accumulation in Appendix A.2 of the supplement.
(For completeness, we confirmed that our implementation of prototypical networks reproduces reported results at higher way.)
}.

\begin{table*}[t]
\caption{
Fully-supervised few-shot accuracy using 100$\%$ of the labeled data.
IMP performs equal to or better than prototypical networks \citep{snell2017prototypical}.
Although IMP is more general, it can still recover uni-modal clustering as a special case.
}
\label{tab:full}
\vskip 0.15in
\begin{center}
\begin{small}
\begin{sc}
\begin{tabular}{lllllll}
& \multicolumn{4}{c}{{\bf Omniglot}} & \multicolumn{2}{c}{{\bf mini-ImageNet}} \\
  & \multicolumn{2}{c}{5-way} & \multicolumn{2}{c}{20-way} & \multicolumn{2}{c}{5-way} \\
{\bf Method}   & 1-shot  & 5-shot & 1-shot & 5-shot & 1-shot & 5-shot \\
\midrule
IMP (ours) & 98.4$\pm$0.3 & 99.5$\pm$0.1 & 95.0$\pm$0.1 & 98.6$\pm$0.1 & 49.6$\pm$0.8 & \textbf{68.1$\pm$0.8} \\
Neighbors & 98.4$\pm$0.3 & 99.4$\pm$0.1 & 95.0$\pm$0.1 & 98.3$\pm$0.1& 49.6$\pm$0.8 & 59.4$\pm$1.0 \\
\citet{snell2017prototypical} & 98.2$\pm$0.3 & 99.6$\pm$0.1 & 94.9$\pm$0.2 & 98.6$\pm$0.1 & 47.0$\pm$0.8 & 66.1$\pm$0.7 \\
\citet{finn2017model} & 98.7$\pm$0.4 & 99.9$\pm$0.3 & 95.8$\pm$0.3 & \textbf{98.9}$\pm$0.2 &  48.7$\pm$1.84 & 63.1$\pm$0.92 \\
\citet{garcia2017few} & \textbf{99.2} & 99.7 & \textbf{97.4} & \textbf{99} & \textbf{50.3} & 66.41 \\
\citet{KaiserNRB17} & 98.4 & 99.6 & 95 & 98.6 & - & - \\
\bottomrule
\end{tabular}
\end{sc}
\end{small}
\end{center}
\vskip -0.1in
\end{table*}

\begin{table*}[t]
\caption{
Semi-supervised few-shot accuracy on $40\%$ of the labeled data with 5 unlabeled examples per class and 5 distractor classes.
The distractor classes are drawn from the complement of the support classes and are only included unlabeled.
IMP achieves equal or better accuracy than semi-supervised prototypical networks \citep{ren18fewshotssl}.
}
\label{tab:semi}
\vskip 0.15in
\begin{center}
\begin{small}
\begin{sc}
\begin{tabular}{lllllll}
& \multicolumn{4}{c}{{\bf Omniglot}} & \multicolumn{2}{c}{{\bf mini-ImageNet}} \\
  & \multicolumn{2}{c}{5-way} & \multicolumn{2}{c}{20-way} & \multicolumn{2}{c}{5-way} \\
{\bf Method}   & 1-shot  & 5-shot & 1-shot & 5-shot & 1-shot & 5-shot \\
\midrule
IMP (ours) & \textbf{98.9 $\pm$ 0.1} & \textbf{99.4 $\pm$ 0.1} & \textbf{96.9 $\pm$ 0.2} & \textbf{98.3 $\pm$ 0.1} & \textbf{49.2 $\pm$ 0.7} & \textbf{64.7 $\pm$ 0.7}  \\
\citet{ren18fewshotssl} & 98.0 $\pm$ 0.1  & \textbf{99.3 $\pm$ 0.1} & 96.2 $\pm$ 0.1 & 98.2 $\pm$ 0.1 & \textbf{48.6 $\pm$ 0.6}  & 63.0 $\pm$ 0.8\\
\midrule
Neighbors & 97.9 $\pm$ 0.2& 99.1 $\pm$ 0.1& 93.8 $\pm$ 0.2& 97.5 $\pm$ 0.1& 47.9 $\pm$ 0.7 & 57.3 $\pm$ 0.8\\
\citet{snell2017prototypical} & 97.8 $\pm$ 0.1 & 99.2 $\pm$ 0.1 & 93.4 $\pm$ 0.1 & 98.1 $\pm$ 0.1 & 45.1 $\pm$ 1.0 & 62.5 $\pm$ 0.5 \\
\bottomrule
\end{tabular}
\end{sc}
\end{small}
\end{center}
\vskip -0.1in
\end{table*}

%% file: rw.tex
\section{Related Work}

\minisection{Prototypes}
Prototypical networks \citep{snell2017prototypical} and semi-supervised prototypical networks \citep{ren18fewshotssl} are the most closely related to our work.
Prototypical networks simply and efficiently represent each class by its mean in a learned embedding.
They assume that the data is fully labeled and uni-modal in the embedding.
\citet{ren18fewshotssl} extend prototypes to the semi-supervised setting by refining prototypes through soft $k$-means clustering of the unlabeled data.
They assume that the data is at least partially labeled and retain the uni-modality assumption.
Both \citet{snell2017prototypical} and \citet{ren18fewshotssl} are limited to one cluster per class.
\citet{mensink2013distance} represent classes by the mean of their examples in a linear embedding to incorporate new classes into large-scale classifiers without re-training.
They extend their approach to represent classes by multiple prototypes, but the number of prototypes per class is fixed and hand-tuned, and their approach does not incorporate unlabeled data.
We define a more general and adaptive approach through infinite mixture modeling that extends prototypical networks to multi-modal clustering, with one or many clusters per class, of labeled and unlabeled data alike.

\minisection{Metric Learning} 
Learning a metric to measure a given notion of distance/similarity addresses recognition by retrieval: given an unlabeled example, find the closest labeled example.
\citet{kulis2013metric} gives a general survey.
The contrastive loss and siamese network architecture \citep{chopra2005learning,hadsell2006dimensionality} optimize an embedding for metric learning by pushing similar pairs together and pulling dissimilar pairs apart.
Of particular note is research in face recognition, where a same/different retrieval metric is used for many-way classification \citep{schroff2015facenet}.
Our approach is more aligned with metric learning by meta-learning \citep{koch2015siamese,vinyals2016matching,snell2017prototypical,garcia2017few}.
These approaches learn a distance function by directly optimizing the task loss, such as cross-entropy for classification, through episodic optimization \citep{vinyals2016matching} for each setting of way and shot.
Unlike metric learning on either neighbors \citep{goldberger2004neighbourhood,schroff2015facenet} or prototypes \citep{snell2017prototypical,ren18fewshotssl}, our method adaptively interpolates between neighbor and uni-modal prototype representation by deciding the number of modes during clustering.

\minisection{Cognitive Theories of Categorization}
Our approach is inspired by the study of categorization in cognitive science.
Exemplar theory \citep{nosofsky1986attention} represents a category by storing its examples.
Prototype theory \citep{reed1972pattern} represents a category by summarizing its examples, by for instance taking their mean.
\citet{vanpaemel2005varying} recognize that exemplars and prototypes are two extremes, and define intermediate models that represent a category by several clusters in their varying abstraction model.
However, they do not define how to choose the clusters or their number, nor do they consider representation learning.
\citet{griffiths2007unifying} unify exemplar and prototype categorization through the hierarchical Dirichlet process to model the transition from prototypes to exemplars as more data is collected.
They obtain good fits for human data, but do not consider representation learning.

%% file: conclusion.tex
\section{Conclusion}
We made a case for the importance of considering the complexity of the data distribution in the regime of few-shot learning.
By incorporating infinite mixture modeling with deep metric learning, we developed infinite mixture prototypes, a method capable of adapting its model capacity to the given data.
Our multi-modal extension of prototypical networks additionally allows for fully unsupervised inference, and the natural incorporation of semi-supervised data during learning.
As few-shot learning is applied to increasingly challenging tasks, models with adaptive complexity will become more important.
Future work will look at extending IMP to the life-long setting, as well as integrating multiple input modalities.

%% file: appendix.tex
\section{Appendix}

\subsection{Implementation Details}
For all few-shot experiments, we use the same base architecture as prototypical networks for the embedding network.
It is composed of four convolutional blocks consisting of a 64-filter $3 \times 3$ convolution, a batch normalization layer, a ReLU nonlinearity, and a $2 \times 2$ max-pooling layer per block.
This results in a 64-dimensional embedding vector for omniglot, and a 1600 dimensional embedding vector for mini-imagenet.
Our models were trained via SGD with RMSProp \citep{tieleman2012lecture} with an $\alpha$ parameter of 0.9.

For Omniglot, the initial learning rate was set to 1e-3, and cut by a factor of two every 2,000 iterations, starting at 4,000 iterations.
Optimization is stopped at 160,000 iterations.
We use gradient accumulation and accumulate gradients over eight episodes before making an update when performing 5-way training.
Both $\sigma_l$ and $\sigma_u$ are initialized to $5.0$.
$\sigma_l$ is learned jointly during training while we found learning $\sigma_u$ on Omniglot to be unstable and so it is therefore fixed.
$\alpha$ was set to 0.1.

For mini-ImageNet, the initial learning rate was set to 1e-3, then halved every 20,000 iterations, starting at 20,000 iterations.
Optimization is stopped at 100,000 iterations.
Both $\sigma_u$ and $\sigma_l$ are initialized to 15.0 and both are learned jointly.
We found that on average, $\sigma_l$ stabilized around 12, and $\sigma_u$ stabilized around 25.
$\alpha$ was set to $10^{-5}$. 
Clusters were still regularly created even with such a small $\alpha$.

\begin{table*}[t]
\caption{
Results on Omniglot for different gradient accumulations.
Bolded results are not significantly different from each other, showing that equalizing the number of gradients can equalize the accuracy.
}
\label{tab:grads}
\vskip 0.15in
\begin{center}
\begin{small}
\begin{sc}
\begin{tabular}{lllllll}
& & & \multicolumn{2}{c}{\textbf{5-way}} & \multicolumn{2}{c}{\textbf{20-way}}\\
Shot & Batch-way & Episode-way & 1-shot & 5-shot & 1-shot & 5-shot \\
\midrule
1 & 20 & 20 & \textbf{98.5} & \textbf{99.6} & \textbf{95.0} & \textbf{98.8} \\
1 & 20 & 5 & \textbf{98.3} & \textbf{99.5} & \textbf{94.8} & \textbf{98.6} \\
1 & 5 & 5 & 97.7 & \textbf{99.4} & 92.1 & 98.0 \\
\midrule
5 & 20 & 20 & \textbf{97.8} & \textbf{99.6} & \textbf{93.2} & \textbf{98.6} \\
5 & 20 & 5 & \textbf{97.9} & \textbf{99.6} & \textbf{92.9} & \textbf{98.5}\\
5 & 5 & 5 & 96.8 & \textbf{99.4} & 89.8 & 97.7\\
\bottomrule
\end{tabular}
\end{sc}
\end{small}
\end{center}
\vskip -0.1in
\end{table*}

\subsection{Controlling for the Number of Gradients Taken During Optimization}
\label{sec:grad-control}

Consider the gradient of the loss: it has the dimensions of shot $\times$ way because every example has a derivative with respect to every class.
In this manner, by default, the episode size determines the number of gradients in an update.
Quantitatively, 20-way episodes accumulate 16 times as many gradients as 5-way episodes.
By sampling 16 5-way episodes and accumulating the gradients to make an update, we achieve significantly better results, matching the results obtained with 20-way episodes within statistical significance in most settings.
Note that agreement across conditions may not be perfectly exact because subtle adjustments to hyperparameters might be necessary. 
See Table \ref{tab:grads} for the quantitative results of these control experiments.

\subsection{Alternative Infinite Mixture Model Algorithms}
\label{sec:crp}
Here we discuss two alternatives to IMP for performing inference in infinite mixture models. We will first discuss an approximation to a Gibbs sampler for estimating the MAP of a Chinese restaurant process (CRP) \cite{aldous1985exchangeability}. We will then discuss an expectation maximization procedure which maintains soft assignments throughout inference.

The generative model of the CRP consists of sampling assignments $z_1, ..., z_J$ which could take on cluster values $c=1,...,C$ from the CRP prior with hyperparameter $\alpha$, which controls the concentration of clusters, and number of cluster members $N_c$.
Cluster parameters $\mu_c$ are sampled from a base distribution $H(\theta_0; \mu_0, \sigma_0)$, and instances $x_j$ are then sampled from the associated Gaussian distribution $ \mathcal{N}(\mu_{z_j}, \sigma_{z_j})$.
$\theta$ consists of the means $\mu$ and sigmas $\sigma$.

The CRP generative model is defined as \begin{align*}
p(z_{J+1}=c|z_{1:J},\alpha)=\frac{N_c}{N+\alpha}\text{ for } c \in \{1 \dots C\} \text{ and }\\
p(z_{J+1}=C+1|z_{1:J}, \alpha)=\frac{\alpha}{N + \alpha}
\end{align*}
for assignments $z$ of examples $x$ to clusters $c$, cluster counts $N_c$, and parameter $\alpha$ to control assignments to new clusters.
$N$ is the total number of examples observed so far.

One popular sampling procedure for parameter estimation is Gibbs sampling \citep{neal2000markov}.
In Gibbs sampling, we draw from a conditional distribution on the cluster assignments until convergence. The conditional draws are:
\begin{equation}
\small
    p(z_{J+1}=c| z_{1:J}, \alpha) \propto \begin{cases} N_{c,-j} \int P(x_j|\theta)dH_{-j,c}(\theta) \text{ for $c \leq C$} \\
    \alpha \int P(x_j | \theta)dH_0(\theta) \text{ for $c = C+1$ } \\
\end{cases}
\end{equation}

For the case of a spherical Gaussian likelihood, let us define $\mathcal{N}_c = \mathcal{N}(x_i; \mu_c, \sigma)$ as the likelihood of assigning $x_i$ to cluster $c$ and $\mathcal{N}_0 = \mathcal{N}(x_i;\mu_0, \sigma+\sigma_0) $ as the likelihood of assigning $x_i$ to a new cluster drawn from the base distribution (Gaussian with mean $\mu_0$ and $\sigma_0$) . We can then write:
\begin{align*}
p(z_i = c | \mu) &= \frac{N_{k,-n}\mathcal{N}_c}{\alpha\mathcal{N}_0 + \sum_{j=1}^C{N_{j,-n}\mathcal{N}_j}} \\
p(z_i = C+1 | \mu) &= \frac{\alpha\mathcal{N}_0}{\alpha\mathcal{N}_0 + \sum_{j=1}^{C}N_{j,-n}\mathcal{N}_j} \\
p(\sigma_c|z) &= \frac{\sigma \sigma_0}{\sigma + \sigma_0N_c} \\
p(\mu_c|z) &= \mathcal{N}\left(\mu_c; \frac{\sigma\mu_0 + \sigma_0\sum_{i, z_i=c}{x_i}}{\sigma+\sigma_0N_c}, \sigma_c\right)
\end{align*}

Unfortunately, because inference must be performed during every episode of our learning procedure, and there are many episodes, Gibbs sampling until convergence is impractical. 
We therefore use the approach from \citep{raykov2016simple} to approximate the procedure with a single pass over all data in the episode.
This approximates the MAP by considering only the most probable cluster assignment for each data point, and updating cluster parameters based on these assignments.
A full discussion is given in \citet{raykov2016simple}, and we include their method here for reference (Algorithm \ref{alg:crp-kmeans}).
While their method is fully-unsupervised, we employ a cross-entropy loss on the query points given the updated means and counts for the \emph{labeled} clusters, for end-to-end optimization of classification, and initialize clusters with the class-wise means as in IMP.

Results for 5-way 1-shot Omniglot and mini-ImageNet are in Table \ref{tab:hard-soft}.
Unlabeled points are often incorrectly assigned to the labeled clusters, which both reduces the variance of that cluster, and increases its likelihood via the prior.
The hard assignments lead to unstable clustering, making learning substantially more challenging.

We additionally implemented a simple expectation maximization approach (Algorithm \ref{alg:em-crp}).
Here we maintain soft assignments $z$ throughout, and use the updates to the cluster means $\mu_c$ as in \cite{kimura2013expectation}.
Our three main differences are to: 1. include labeled points for initialization; 2. instead of having a fixed truncation parameter $T$ for the maximum number of available clusters, we instantiate new clusters when the probability of a new cluster exceeds a certain threshold $\epsilon$; 3. we do not estimate variances, as this led to very unstable results.
Instead of estimating variances based on assignments, we use the same variance learning technique as IMP, which provides significant improvement.
The best value of $\alpha$ was one for which no new clusters were created in both Omniglot and mini-ImageNet.

\begin{table}[H]
\caption{Ablation experiments comparing different inference schemes for infinite mixture prototypes. Accuracies are for semi-supervised 5-way 1-shot episodes, with 5 unlabeled examples per class, and 5 distractors.}
\label{tab:hard-soft}
\vskip 0.15in
\begin{center}
\begin{small}
\begin{sc}
\begin{tabular}{lcc}
Method  & Omniglot & mini-ImageNet \\
\midrule
MAP-DP ($\mu,\sigma$) & 70.0 $\pm$ 0.1 & unstable \\
EM & 95.9 $\pm$ 0.2 & 41.0 $\pm$ 0.6\\
Hard DP-means & 98.0 $\pm$ 0.2 & 45.2 $\pm$ 1.0\\
IMP & \textbf{99.0} $\pm$ 0.1 & \textbf{49.6} $\pm$ 0.6\\
\bottomrule
\end{tabular}
\end{sc}
\end{small}
\end{center}
\vskip -0.1in
\end{table}

We additionally tested the hypothesis that the CRP prior was leading to worse performance by ablating it. 
With the prior ablated, the EM approach improves to 48.6$\%$ accuracy on mini-ImageNet, and 98.0$\%$ accuracy on Omniglot. 
While this is still below IMP's performance, this gives some explanation for why the EM inference procedure fails.

The experiments in this section examine the semi-supervised 5-way 1-shot setting, with 5 unlabeled examples of each character and 5 distractor classes (see Section 4.2 of the paper for more experimental detail).
In this setting, there is no effect of multi-modality in the labeled examples, and so any improvements by IMP are attributed to the way it clusters \emph{unlabeled} data relative to these inference methods.

\newpage

\begin{algorithm*}[h]
\caption{MAP-DP approach for inference.
\label{alg:crp-kmeans}
$n_s$ is the number of labeled classes (way).
$q(i,c)$ is $\log p(i, c)$, the joint probability of cluster $C$ and assignment $i$.
$\mathcal{N}(x;\mu,\sigma)$ is the Gaussian density.
$\alpha$ is the concentration hyperparameter of the CRP.
}
\label{MAP}
\begin{algorithmic}
\STATE initialize $\{\mu_1, \dots, \mu_{n_s}\}$ \COMMENT{Initialize a cluster for each labeled class by taking class-wise means}
\STATE initialize $\{\sigma_1, \dots, \sigma_{n_s}\}$ \COMMENT{Initialize cluster variances based on equation 4.}
\STATE initialize $\{z_1, \dots, z_{I}\}$ \COMMENT{Initialize cluster assignments for labeled data points. All unlabeled cluster assignments start at 0.}
\STATE $C = n_s$ \COMMENT{Initialize number of clusters $C$}

\fullcomment{Begin clustering pass}
\FOR{each example $i$}
\FOR{each cluster $c \in \{1,...,C\}$}
    \STATE $N_c \gets \sum_i z_{i,c}$
    \STATE $\sigma_c \gets \frac{\sigma\sigma_0}{\sigma+\sigma_0N_c}$
    \STATE $\mu_c \gets \frac{\sigma\mu_0 + \sigma_0\sum_{i}z_{i,c}{h_{\phi}(x_i)}}{\sigma + \sigma_0N_c}$
	\STATE estimate $q_{i,c} \propto \log(N_{c,-i}) + \log(\mathcal{N}(x_i; \mu_c, \sigma_c))$ 
\ENDFOR
\STATE estimate $q_{i,C+1} \propto \log(\alpha) + \log(\mathcal{N}_0(x_i; \mu_0, \sigma_0))$ 
\STATE $z_{i} \gets argmin(q_{i,1},...,q_{i,C+1})$
\IF{$z_{i} = C+1$}
\STATE $C \gets C+1$
\ENDIF
\ENDFOR
\end{algorithmic}
\end{algorithm*}

\begin{algorithm*}[h]
\caption{EM approach for inference.
\label{alg:em-crp}
$n_s$ is the number of labeled classes (way).
$q(i,c)$ is $\log p(i, c)$, the joint probability of cluster $C$ and assignment $i$.
$\mathcal{N}(x;\mu,\sigma)$ is the Gaussian density.
$\alpha$ is the concentration hyperparameter of the CRP.
$\epsilon$ threshold for generating new cluster.
}
\label{MAP}
\begin{algorithmic}
\STATE initialize $\{\mu_1, \dots, \mu_{n_s}\}$ \COMMENT{Initialize a cluster for each labeled class by taking class-wise means}
\STATE initialize $\{\sigma_1, \dots, \sigma_{n_s}\}$ \COMMENT{Initialize cluster variances based on equation 4.}
\STATE initialize $\{z_1, \dots, z_{I}\}$ \COMMENT{Initialize cluster assignments for labeled data points. All unlabeled cluster assignments start at 0.}
\STATE $C = n_s$ \COMMENT{Initialize number of clusters $C$}

\fullcomment{Begin clustering pass}
\FOR{each example $i$}
\FOR{each cluster $c \in \{1,...,C\}$}
	\STATE estimate $q_{i,c} \propto \log(N_{c,-i}) + \log(\mathcal{N}(x_i; \mu_c, \sigma_c))$ 
\ENDFOR
\STATE estimate $q_{i,C+1} \propto \log(\alpha) + \log(\mathcal{N}_0(x_i; \mu_0, \sigma_0))$ 
\STATE $z_{i,c} \gets $softmax$(q_{i,1},...,q_{i,C+1})$
\IF{$z_{i,C+1} > \epsilon$}
\STATE $C \gets C+1$
\STATE $\mu_{C} \sim \mathcal{N}(x_i, \mu_0, \sigma_0)$ \COMMENT{Sample from the base distribution conditioned on the single observation $x_i$}
\ENDIF
\ENDFOR
\end{algorithmic}
\end{algorithm*}